\documentclass[letterpaper, 10 pt, conference]{ieeeconf}  

\IEEEoverridecommandlockouts                              

\makeatletter
\let\NAT@parse\undefined
\makeatother

\usepackage{times}
\usepackage{epsfig}
\usepackage{graphicx}
\usepackage{amsmath}
\usepackage{amssymb}
\usepackage{booktabs}
\usepackage{array}
\usepackage[dvipsnames]{xcolor}
\usepackage{multirow}
\usepackage{tabularx}
\usepackage{siunitx}
\usepackage{cprotect}
\usepackage{svg}
\usepackage{stfloats}
\usepackage{afterpage}
\usepackage[colorlinks]{hyperref}
\usepackage{cite}
\usepackage{subcaption}
\renewcommand{\baselinestretch}{0.98} 

\title{\LARGE \bf
AERO-VIS:\\Asynchronous Event-based Real-time Onboard Visual-Inertial SLAM
}

\author{Yannick Burkhardt$^{1,2,3}$, Sebastián Barbas Laina$^{1,2,4}$, Simon Boche$^{1,2}$, Leonard Freissmuth$^{1}$, Stefan Leutenegger$^{1}$
\thanks{$^{1}$Mobile Robotics Lab, ETH Zürich, \{\href{mailto:yannick.burkhardt@mrl.ethz.ch}{\nolinkurl{yannick.burkhardt}}, \hspace{0.5pt}\href{mailto:sebastian.laina@mrl.ethz.ch}{\nolinkurl{sebastian.laina}}, \hspace{0.5pt}\href{mailto:simon.boche@mrl.ethz.ch}{\nolinkurl{simon.boche}}, \href{mailto:leonard.freissmuth@mrl.ethz.ch}{\nolinkurl{leonard.freissmuth}}, \hspace{0.5pt}\href{mailto:stefan.leutenegger@mrl.ethz.ch}{\nolinkurl{stefan.leutenegger}}\}\href{mrl.ethz.ch}{\nolinkurl{@mrl.ethz.ch}} }%
\thanks{$^{2}$Technical University of Munich}%
\thanks{$^{3}$Munich Center for Machine Lerning (MCML)}%
\thanks{$^{4}$Munich Institute of Robotics and Machine Intelligence (MIRMI)\newline
Accepted for publication in IEEE Robotics and Automation Letters (RA-L), August 2026.}%
}

\makeatletter
\def\@IEEEauthorblockconfadjspace{-1.0em}   
\makeatother
\IEEEaftertitletext{\vspace{-1\baselineskip}}  

\begin{document}

\maketitle
\thispagestyle{empty}
\pagestyle{empty}

\begin{abstract}
The robustness of event cameras to high dynamic range and motion blur holds the potential to improve visual odometry systems in challenging environments. Although their high temporal resolution does not require synchronous processing, most event-based odometry methods still run at fixed rates, which simplifies system design but restricts latency and throughput. In this work, we present AERO-VIS, a stereo event-inertial SLAM system with an integrated, data-driven, robust, and performance-optimized keypoint detector. By decoupling event preprocessing from state estimation, our architecture operates asynchronously at the maximum rate supported by available hardware, dynamically adapting to downstream runtime demands and ensuring low-latency, real-time performance. When deploying AERO-VIS on a UAV, we achieve unprecedented accuracy in onboard event-based SLAM. These unique characteristics enable us to present the first purely event-based inertial SLAM system that demonstrates closed-loop UAV control and large-scale state estimation while relying solely on onboard compute. Videos of the experiments, source code, and additional results are available at \href{https://ethz-mrl.github.io/AERO-VIS/}{\nolinkurl{ethz-mrl.github.io/AERO-VIS}}.
\end{abstract}
\afterpage{
\begin{figure}[t]
    \centering
    \begin{subfigure}{\columnwidth}
        \centering
        \includegraphics[width=\columnwidth]{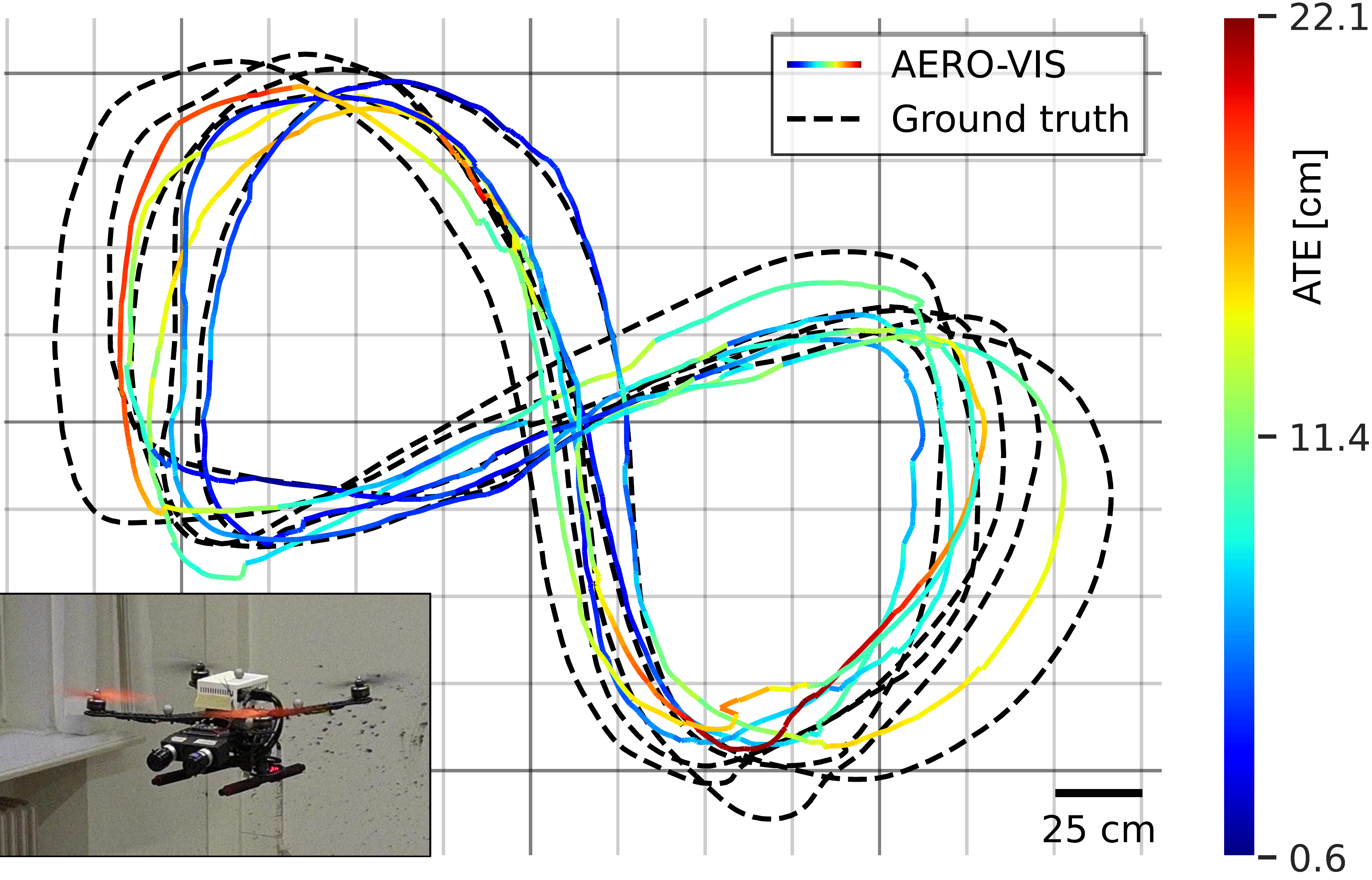}
        \vspace{-16pt}
        \caption{Closed-loop control under normal conditions.}
        \vspace{4pt}
    \end{subfigure}
    \begin{subfigure}{0.45\columnwidth}
        \includegraphics[width=\columnwidth]{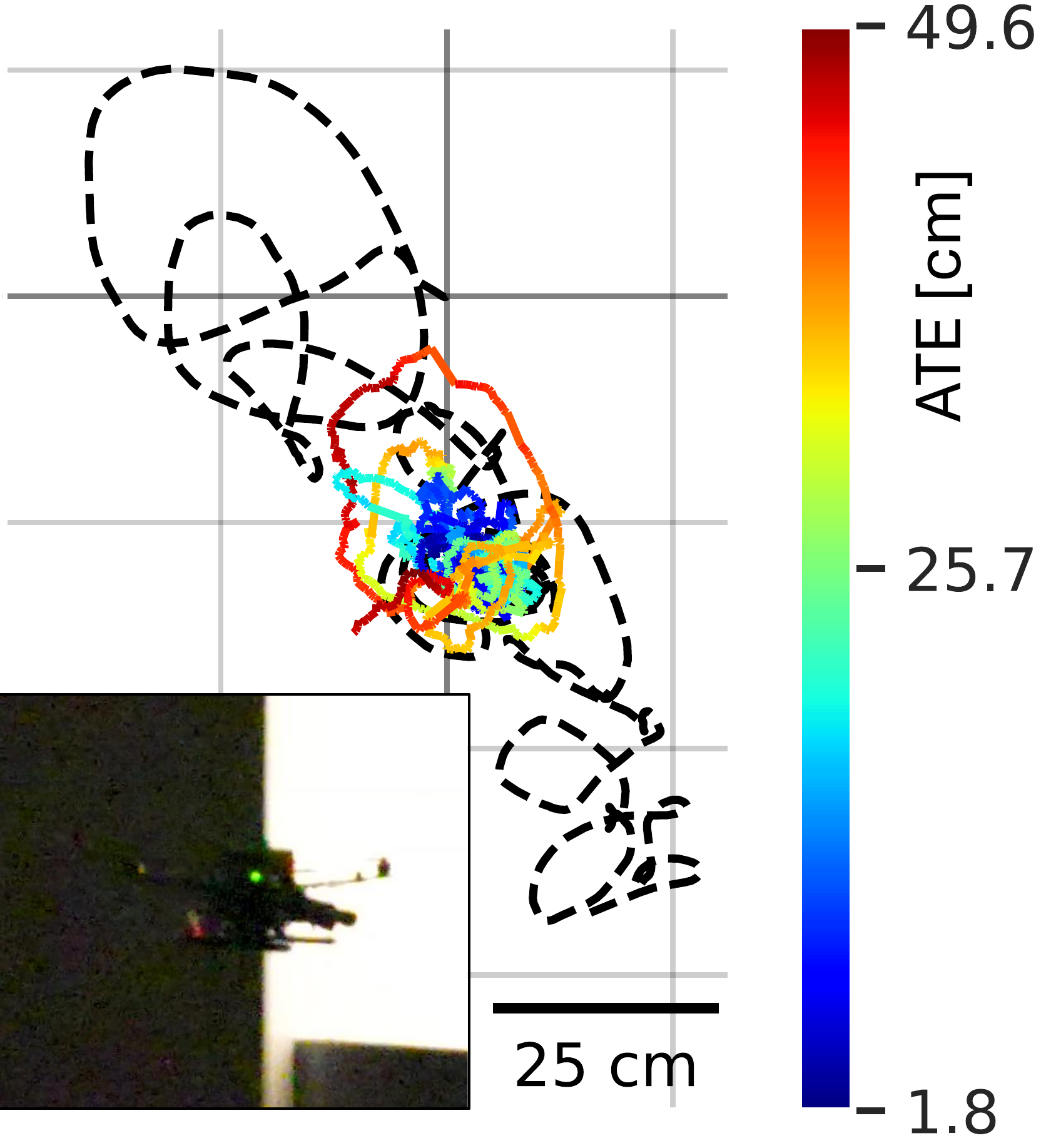}
        \vspace{-16pt}
        \caption{Closed-loop control in HDR where frame-based OKVIS2 fails.}
    \end{subfigure}\hfill
    \begin{subfigure}{0.51\columnwidth}
        \includegraphics[width=\columnwidth]{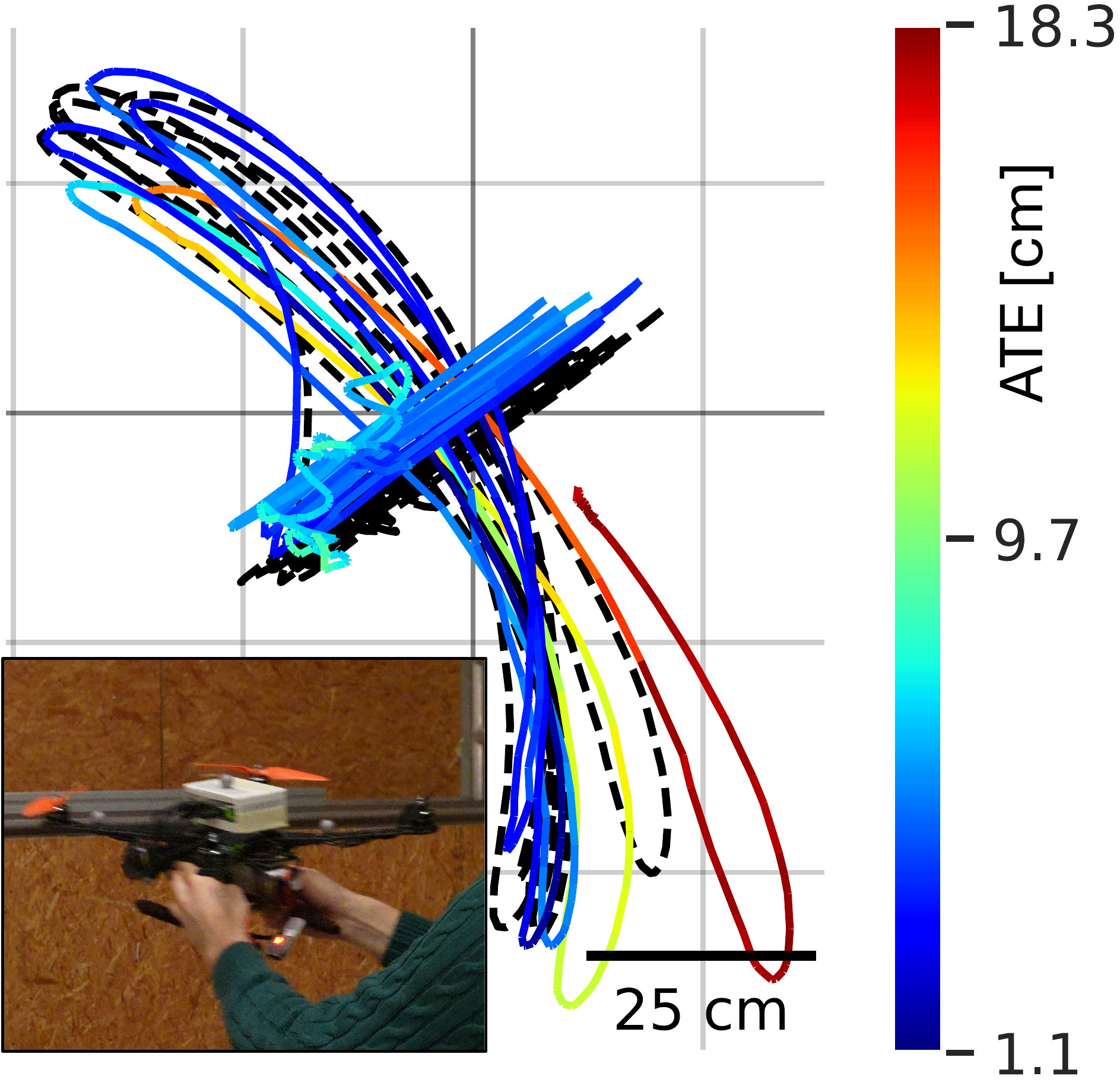}
        \vspace{-16pt}
        \caption{Aggressive handheld motion with velocities up to \SI{3}{m \per s} and \SI{270}{\degree \per s}.}
    \end{subfigure}
    \caption{Estimated trajectories by AERO-VIS when employed onboard a UAV.}
    \vspace{-8pt}
  \label{drone_traj}
\end{figure}
}

\section{Introduction}
Visual odometry (VO) and SLAM systems provide the essential localization and mapping capabilities required for fully autonomous robotic navigation. However, the reliability of these systems can be compromised by the inherent limitations of traditional frame-based sensors in high-speed or low-light environments. By providing high dynamic range (HDR) and microsecond-level temporal precision resistant to motion blur, event cameras enhance robotic perception in extreme conditions where traditional frame cameras often fail. However, the transition from synchronous frames to asynchronous, sparse event streams demands the development of entirely new algorithmic frameworks capable of handling non-traditional data. The lack of efficient event data processing paradigms forces most current event-based visual VO and SLAM systems to choose between high accuracy and real-time performance. These developments often led to a dependence on large power-hungry computational resources in current state-of-the-art (SOTA) systems~\cite{Bur25, Niu24_b, Zho25} -- a requirement that cannot be satisfied on most mobile robot platforms, such as unmanned aerial vehicles (UAVs).\par
A few systems have been evaluated onboard UAVs~\cite{Che23, Tej26}, but these fall short in terms of accuracy and stability. Consequently, they remain insufficient for the rigorous demands of closed-loop control and high-level autonomy in real-world scenarios. The high event rate triggered by drone vibrations and aggressive motion poses an additional challenge for onboard systems with a limited real-time budget. With this paper, we therefore aim to answer the research question: \textit{Can a UAV be successfully controlled relying on event cameras as the only visual sensing modality?}\par
Addressing this challenge requires an event-based system capable of accurate and efficient pose estimation. Therefore, we modify the SOTA frame-based visual-inertial (VI) SLAM system OKVIS2~\cite{Leu22} and employ a data-driven approach to detect and describe keypoints in the event stream, inspired by~\cite{Bur25}. By improving the Multi-Channel Time Surface (MCTS) event representation and streamlining the network architecture, our model enhances tracking accuracy while significantly reducing computational requirements. To minimize latency while maximizing throughput, we decouple the synchronous, high-rate event retrieval and preprocessing thread from state estimation, which runs asynchronously at the maximum rate supported by the computational resources.\par
Our stereo event-inertial SLAM system, AERO-VIS, achieves superior estimation accuracy on public datasets while running in real-time on an NVIDIA Jetson Orin NX, outperforming state-of-the-art approaches~\cite{Niu24_b, Zho25} without computational constraints. In several real-world experiments, we further validate AERO-VIS's precision and stability. We demonstrate that our system retains the event camera's inherent robustness to HDR illumination and its immunity to motion blur during aggressive drone movement, as visualized in Fig.~\ref{drone_traj}. Additionally, it preserves the robust large-scale trajectory estimation and loop-closure capabilities of OKVIS2.\par

\textbf{Contributions:}
(1) We present AERO-VIS, an event-based SLAM system designed to maximize processing frequency and minimize latency through a multi-threaded, asynchronous architecture, extending the OKVIS2~\cite{Leu22} framework.
(2) We introduce a refined event representation and a streamlined architecture for a SOTA~\cite{Bur25} keypoint detector and descriptor, simultaneously improving its accuracy and achieving a ten-fold reduction in its onboard inference time.
(3) Extensive experiments demonstrate AERO-VIS's robustness and efficiency across multiple public datasets and in real-world scenarios with fast motion, high dynamic range, or large scale. To our knowledge, this work marks the first successful demonstration of closed-loop UAV control with a purely event-inertial SLAM system.
\section{Related Work}
In this section, we review event-based VO and SLAM approaches grouped in feature-based, direct, and data-driven methods, with a special focus on their real-time capability.\par

\textbf{Feature-based methods} infer camera motion by matching keypoints (e.g., corners) across time. The early monocular EVIO~\cite{Zhu17} tracks features in event-frames by compensating for camera motion with local optical flow, subsequently fusing estimates via an extended Kalman filter. Since runtime scales linearly with feature count, the system becomes inefficient during high-density tracking required for stable estimation. Another monocular approach~\cite{Had21} estimates the camera pose change by matching time-surface~\cite{Lag17} patches around Arc* corners~\cite{Alz18} via cross-correlation. However, the motion-dependent features cause instability during motion changes, while Arc*’s event-by-event processing creates computational overhead that limits real-time onboard use. By leveraging line features, the monocular system~\cite{Cha22} employs EMVS~\cite{Reb18} for mapping and a Lie-formulated error-state Kalman filter for pose estimation. It achieves high processing rates but limited robustness and accuracy.\par
ESVIO~\cite{Che23} utilizes a stereo event-inertial setup, with optional support for stereo frame cameras. Building on its monocular predecessor~\cite{Gua22}, the system leverages gyroscope data to compensate for rotational motion, aligning events in the image plane. Arc*-detected corners are tracked with the Lukas-Kanade method~\cite{Kan81} and fused with frame-based matches. While the full ESVIO pipeline -- integrating both frame and event data -- is validated through real-time, closed-loop UAV control, the event-only configuration lacks similar hardware validation. Its inferior performance on standard datasets compared to other event-based solutions~\cite{Niu24_b, Zho25} suggests that ESVIO’s stability depends on the inclusion of frame-based input rather than the event stream alone.\par

\textbf{Direct methods} infer the camera motion by aligning events densely. The first work to achieve 6D event camera pose tracking~\cite{Kim16} estimates log-intensity images to estimate depth and camera motion under the constant-brightness assumption. It leverages a GPU to achieve real-time estimation with low-resolution event cameras. EVO~\cite{Reb16} increases efficiency by estimating camera motion and a semi-dense map through geometric alignment of binary event frames. However, both monocular approaches show limited accuracy and robustness compared to modern methods~\cite{Niu24_b, Zho25}.\par
ES-PTAM~\cite{Gho24} employs MC-EMVS~\cite{Gho22}, a multi-camera mapping framework that optimizes the ray reprojection of events observed from different camera views, along with edge-map alignment for pose estimation. In addition to significant drift in the evaluation results, the system requires the ground truth pose for initialization, severely limiting its usability in real-world scenarios.\par
The stereo event-based system ESVO~\cite{Zho21} optimizes the spatio-temporal consistency of time surfaces for localization and mapping. While one work extends ESVO with truncated signed distance functions for mapping~\cite{Liu23_t}, others incorporate IMU measurements: \cite{Liu23} improves event alignment through motion compensation, while~\cite{Niu24_a, Niu24_b} introduce tightly coupled event-inertial optimization and a refined event representation based on edge-pixel sampling. The resulting system, ESVO2~\cite{Niu24_b}, achieves SOTA efficiency, processing VGA-resolution event streams in real-time on a desktop PC.\par

\textbf{Data-driven methods} employ deep learning to obtain a statistical model capable of predicting the camera pose (end-to-end) or to tackle a sub-problem, e.g., keypoint detection (hybrid). Some self-supervised end-to-end learning approaches predict monocular optical flow, depth, and egomotion~\cite{Ye18, Zhu19} but struggle to generalize beyond the specific driving scenarios encountered during training.\par
DEVO~\cite{Kle24} attempts to overcome this limitation by adapting the monocular, frame-based end-to-end model DPVO~\cite{Tee23} to the event domain, introducing a patch selection mechanism and training on a large synthetic dataset. The promising results suffer from scale ambiguity, an inherent limitation of monocular systems. Two approaches address this problem by extending DEVO to additional sensors. DEIO~\cite{Gua24_b} additionally incorporates inertial measurements, however, its metric-scale estimation remains unstable. Rather than relying on an IMU, SDEVO~\cite{Zho25} employs a novel static stereo association method to recover metric scale through stereo event cameras. While both DEVO and DEIO are not real-time optimized, SDEVO streamlines data processing and employs a mixed-precision model to operate at \SI{15}{Hz} at VGA resolution on a desktop PC with a modern GPU.\par
A hybrid monocular VIO system~\cite{Tej26} reconstructs frames from the event stream with the efficient, data-driven FireNet~\cite{Sch20} and processes them with OpenVINS~\cite{Gen20}. Its lightweight components allow onboard pose estimation at \SI{20}{Hz}. However, frame reconstruction can produce artifacts, introduces computational overhead and neutralizes the high temporal resolution of event cameras.\par
A demonstration of a spiking neural network mapping raw monocular event data directly to control commands for hovering, landing, and lateral flight~\cite{Par24} relies on a static ground-plane assumption that restricts it to set-point-following over flat surfaces, and processes only events within corner patches, as data-driven full-stream event-by-event processing~\cite{Sek26, Hao26} is not yet real-time capable on embedded hardware.\par
Recently, the data-driven method SuperEvent~\cite{Bur25} emerged to improve the robustness of event-based keypoint detection and description. It adapts SuperPoint~\cite{Det18} to the event domain by employing the MCTS event representation, a multi-channel generalization of time surfaces to mitigate their dependence on scene motion. The authors achieve promising event-based benchmarking results by replacing the frame-based detector of the VI-SLAM system OKVIS2~\cite{Leu22} with SuperEvent. However, it lacks real-time capability and requires a fixed, predefined processing rate. In this work, we build on this approach by improving SuperEvent's accuracy and performance, streamlining data flow, and adapting the OKVIS2 framework for real-time 6-DoF state estimation to achieve real-world onboard estimation in unstructured, large-scale environments, without scene-planarity assumptions.

\section{Method}~\label{sec:method}
AERO-VIS introduces several key innovations. As the system relies on keypoint matching, it necessitates a highly accurate and efficient method for event-based keypoint detection and description. To meet this requirement, we leverage SuperEvent~\cite{Bur25}, based on an enhanced input representation as detailed in Sec.~\ref{sec:mcts}. Furthermore, to accommodate the limited computational resources of embedded hardware, we introduce the runtime-optimized network architecture of SuperLitE in Sec.~\ref{sec:arch} before presenting the asynchronous AERO-VIS framework in Sec.~\ref{sec:system}.

\subsection{Multi-Channel Time Surfaces with Constant Event Count}\label{sec:mcts}
MCTSs generalize time surfaces to more than one channel and have been shown to improve the prediction stability of keypoint detection networks~\cite{Bur25}. A key limitation of traditional time surfaces is their sensitivity to the time window parameter, as its ideal setting varies with the local motion of individual keypoints. To enrich the information for the neural network, the MCTS stacks polarity-separated time surfaces for $K$ different time window durations $\Delta t_k$. These define the $2K$ sets of events
$\mathcal{E}_{p,  \Delta t_k} = \{ \mathbf{e}_i \mid p_i = p, t_i \in [\tau - \Delta t_k, \tau] \}$ at time $\tau$ with each event $\mathbf{e}_i = (t_i, x_i, y_i, p_i)$ characterized by its timestamp $t_i$, pixel coordinates $(x_i,y_i)$, and polarity $p_i \in \{-1,+1\}$. The MCTS tensor at $\tau$ is computed as
\begin{multline}
    \text{\textbf{MCTS}} =
    (\text{\textbf{TS}}_{{-1}, \Delta t_1}, \ldots, \text{\textbf{TS}}_{{-1}, \Delta t_K},\\ \text{\textbf{TS}}_{{+1}, \Delta t_1}, \ldots, \text{\textbf{TS}}_{{+1}, \Delta t_K }),
\end{multline}
with the zero-initialized time surfaces TS
\begin{equation}
   \text{\textbf{TS}}_{{p}, \Delta t_k}(x_i, y_i) = \max_{\mathbf{e}_i \in \mathcal{E}_{p, \Delta t_k}}\left(1 - \frac{\tau - t_i}{\Delta t_k}\right).
\end{equation}

While SuperEvent shows promising results in keypoint detection and description, the event stream's motion dependence limits performance for strong motion changes. Even though the motion-dependent appearance of features is an inherent property of the event camera, the described problem can be mitigated by reformulating the MCTS calculation. Instead of using fixed time windows $\Delta t_k$, we propose to adapt $\Delta t_k$ to obtain a constant number of events $N_{e,k}$ per channel pair $k$
\begin{equation}
    \Delta t_k = \tau - t_{I-N_{e,k}},
\end{equation}
with index $I$ for the most recent event. This results in visually more similar time surfaces for varying motion magnitudes, improving keypoint predictions as shown in Sec.~\ref{sec:se}. We denote the MCTS formulation introduced in SuperEvent~\cite{Bur25} as MCTS$_{\Delta t}$ and our proposed version as MCTS$_{Ne}$, reflecting the constant property. We select the $N_{e,k}$ on the logarithmic ladder of the MCTS$_{\Delta t}$ time windows. As the smallest-count channels attain the lowest first-layer weight magnitudes, removing them (yielding an eight-channel input tensor) preserves accuracy after retraining. To ensure sensor-agnosticism, we normalize the event counts $N_{e,k}$ by the total number of sensor pixels as the event rate scales with spatial resolution. This yields normalized event counts $\overline{N}_{e,k} \in \{0.03, 0.1, 0.3, 1.0\}$ \SI{}{events \per pixel}. In the following, \textit{SuperEvent+} refers to this refined approach, which maintains SuperEvent's architecture with the first layer adjusted to process an eight-channel MCTS$_{Ne}$ input tensor.

\begin{figure*}[t]
  \includegraphics[width=\textwidth]{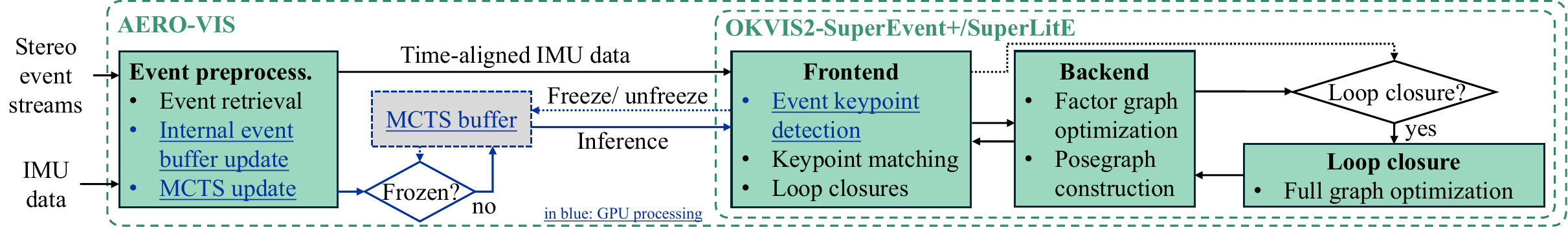}
  \caption{Stereo event and IMU data are preprocessed through time synchronization and high-frequency MCTS calculation in a shared buffer. The frontend processes this buffer asynchronously, freezing it for SuperLitE inference directly after completing the previous iteration's keypoint matching and loop closure detection. The backend optimizes the factor graph of recent sensor observations and constructs the posegraph on every new keyframe. If a loop closure is detected, a detached thread is initiated to re-optimize the global pose graph in the background.}
  \label{system}
  \vspace{-8pt}
\end{figure*}

\subsection{SuperLitE Architecture}\label{sec:arch}
Real-time performance on embedded hardware requires a lightweight network architecture. Our analysis revealed that the shared encoder and descriptor processing constitute the primary computational bottlenecks within the SuperEvent~\cite{Bur25} architecture. We conducted a design space search varying the encoder architectures~\cite{Tu22, Sim14, How19}, network depth, and descriptor dimensionality. We find that an encoder with four layers, each with one convolution, maximum-pooling, and batch-normalization, maintains good accuracy while drastically accelerating onboard inference. Additionally, reducing the descriptor dimensionality from 256 to 64 facilitates faster interpolation of the spatially-reduced descriptor map and accelerates data transfer and matching within our SLAM system. We name the resulting model -- the best accuracy/performance trade-off we found -- \textit{SuperLitE}, where the trailing ``E'' emphasizes its \textit{Event} processing capability. Further reducing the MCTS channel count, descriptor dimension, or backbone depth significantly degrades accuracy.

\subsection{AERO-VIS System Design}\label{sec:system}
OKVIS2~\cite{Leu22} employs keypoint matching for landmark triangulation and residual computation within the frontend thread, while the backend utilizes factor graph optimization to jointly estimate camera poses and landmark positions. By comparing the current set of descriptors with previous keyframes encoded in a DBoW2~\cite{Gal12} database, the frontend detects loop closures. To ensure robustness against false positives, heuristic checks verify that drift remains within realistic bounds before triggering a full factor graph optimization in a background thread.\par
To achieve event-inertial SLAM, we modify the OKVIS2 framework by substituting its conventional frame-based keypoint detector with one of our event-based approaches. Depending on available computational resources and runtime constraints, the system can be configured to use either the full SuperEvent+ model or its runtime-optimized variant, SuperLitE. Additionally, we adjust the calculation of the descriptor distance. Instead of the Euclidean distance employed in~\cite{Bur25}, we use the cosine distance, as it reduces the required computations per matching attempt.\par
In practice, the system's per-iteration processing time varies with operational factors such as the number of observed keypoints and the growing keyframe database. Additionally, the near time-continuous event stream allows state estimation at any given timestamp by generating and processing the corresponding MCTS$_{Ne}$. Exploiting this property, our idea is to process the latest possible event state when the frontend is ready. This reduces the system's latency compared to synchronous (e.g., frame-based) processing.\par
On devices with limited computational resources, creating an MCTS$_{Ne}$ comes with significant computational cost. To avoid unnecessarily reducing the system's throughput, we decouple event preprocessing from state estimation. For multi-modal alignment, only events with corresponding IMU data are processed, while newer events are retained in a buffer. These synchronized events update a two-channel GPU-tensor, storing the most recent event timestamps per pixel and polarity. This tensor then serves as the basis for calculating the $MCTS_{Ne}$ within shared memory. The frontend thread gates this memory, permitting updates only when the memory is not utilized for inference.\par
The preprocessing thread operates at the event sensor's fixed, high-frequency event stream update rate to ensure minimal frontend latency. To maximize throughput, the thread decoupling allows the remaining SLAM system to run asynchronously. Once the frontend thread finished the previous iteration, it freezes the shared MCTS buffer to run SuperLitE inference, and then unfreezes it again. After keypoint selection, the normalized descriptors are quantized from floating-point to 8-bit integers for more efficient copying and calculations. We apply symmetric min-max quantization using a scaling factor derived from the training data. This factor is neutralized during cosine distance calculation.\par
A block diagram of our system \textit{AERO-VIS} can be found in Fig.~\ref{system}. The system is real-time capable by design, since its asynchronous processing rate scales according to the current computational load. Additionally, latency is minimized by always updating the MCTS buffer to the latest state where state estimation is possible.

\section{Experiments}
We conduct various experiments to validate our systems. Firstly, we ablate our proposed event representation and streamlined network architecture for the keypoint-based pose estimation in Sec.~\ref{sec:se}. We then evaluate OKVIS2-SuperEvent+, AERO-VIS and two baselines on three popular benchmarks in Sec.~\ref{sec:data}. We compare nominal accuracy as well as real-time results on an embedded device. To further demonstrate the capabilities of AERO-VIS, we conduct two more real-world experiments. In Sec.~\ref{sec:drone_exp}, we employ our system onboard a UAV with different environmental conditions to achieve closed-loop trajectory following. Finally, we show the system's robustness in a large-scale handheld experiment and analyze the computation time in Sec.~\ref{sec:ls}.

\afterpage{
\begin{figure}[t]
  \includegraphics[width=\columnwidth]{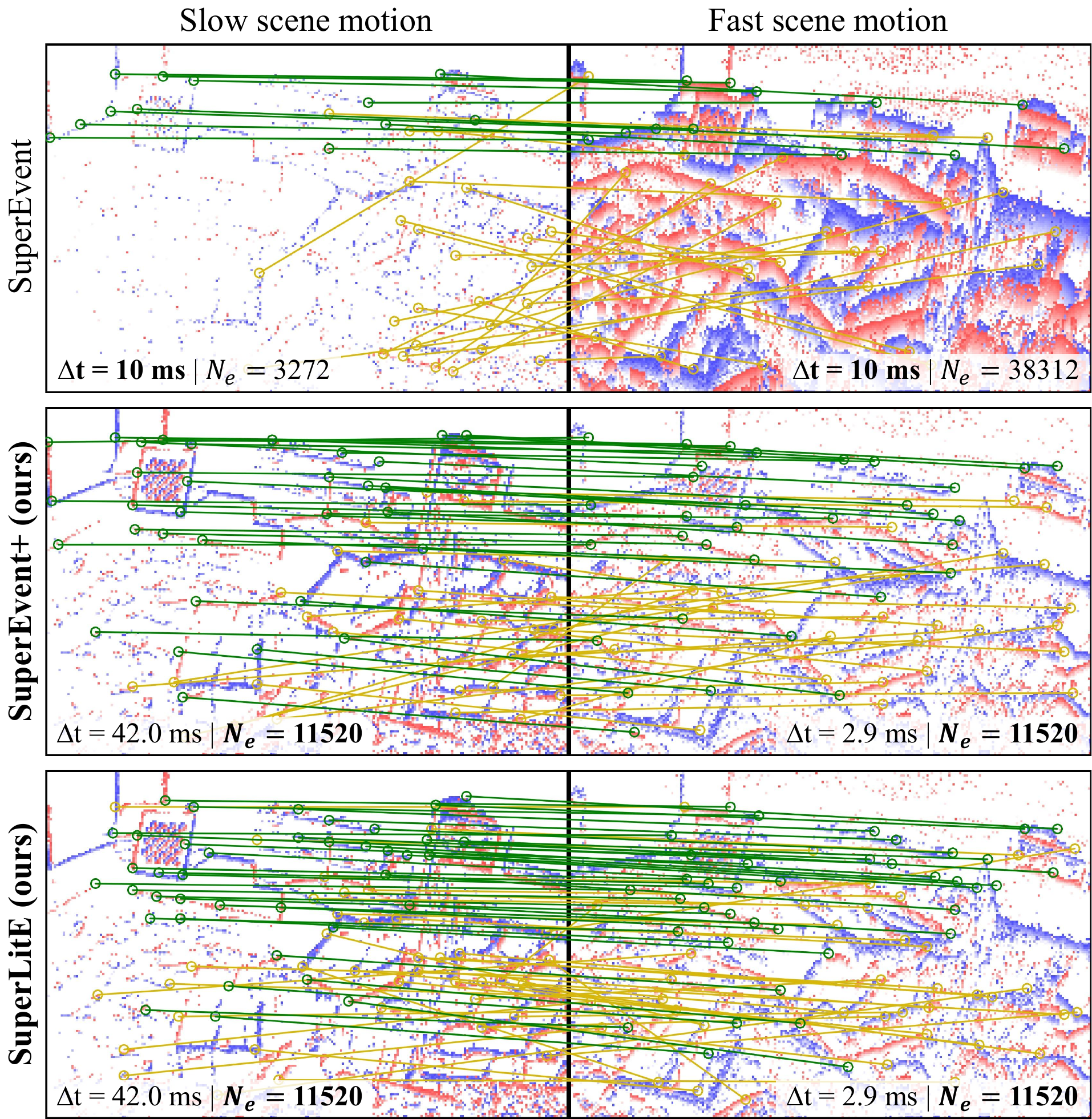}
  \caption{Two samples of the \textit{boxes\_6dof}-sequence of the ECD dataset: we visualize one channel-pair with positive (red) and negative (blue) polarity for MCTS$_{\Delta t}$ (fixed time windows) and MCTS$_{Ne}$ (fixed event counts), as well as the matches predicted by SuperEvent~\cite{Bur25}, SuperEvent+ and SuperLitE. Green matches have a reprojection error of less than 5 pixels using ground truth pose measurements, while yellow matches are outliers.}
  \label{se_infer}
  \vspace{-4pt}
\end{figure}
}

\afterpage{
\begin{table}[t]
\begin{center}
\caption{Area-under-curve (AUC) of relative pose estimation on the Event Camera Dataset (ECD) and Event-aided Direct Sparse Odometry (EDS) of SuperEvent (SE) and our variants SuperEvent+ (SE+), and SuperLitE (SL). Inference times of the captured CUDA graph of the TensorRT-compiled models with 16-bit floating-point precision and $240 \times 400$ pixels input resolution are measured on an NVIDIA Jetson Orin NX.\label{pose_est}}
\begin{tabularx}{\linewidth}{
l
>{\centering}X
>{\centering}X
>{\centering}X
>{\centering}X
>{\centering}X
>{\centering\arraybackslash}X
r
}
\toprule
\multirow{2.75}{*}{Method} & \multicolumn{3}{c}{ECD: AUC [\%]} & \multicolumn{3}{c}{EDS: AUC [\%]} & \multirow{1.75}{*}{Infer.} \\
\cmidrule(lr){2-4}
\cmidrule(lr){5-7}
& @5° & @10° & @20° & @5° & @10° & @20° & time \\
\midrule
SE \cite{Bur25} & \underline{22.7} & 35.8 & 46.7 & 25.3 & 37.5 & 48.9 & 26.2\hspace{2pt}ms \\

\textbf{SE+ (ours)} & \textbf{24.3} & \textbf{40.1} & \textbf{53.3} & \textbf{32.4} & \textbf{47.0} & \textbf{60.2} & \underline{25.1\hspace{2pt}ms} \\

\textbf{SL (ours}) & 22.3 & \underline{38.4} & \underline{52.8} & \underline{31.2} & \underline{46.0} & \underline{59.3} & \textbf{2.5\hspace{2pt}ms} \\

\bottomrule
\vspace{-6pt}
\end{tabularx}
\vspace{-18pt}
\end{center}
\end{table}
}

\subsection{SuperEvent+ \& SuperLitE}~\label{sec:se}
We adopt the training strategy and relative pose estimation evaluation from SuperEvent~\cite{Bur25}. The area-under-curve (AUC) scores for different rotation error thresholds, along with inference times on an embedded device, are reported in Table~\ref{pose_est}. We benchmark on the Event Camera Dataset~\cite{Mue17_ecd} captured with an iniVation DAVIS240C event camera at a $180\times240$ pixel resolution, and the Event-aided Direct Sparse Odometry dataset (EDS)~\cite{Hid22} recorded with a Prophesee Gen 3.1 event camera with $480\times640$ pixel resolution.\par
SuperEvent+ achieves an average relative improvement of 18.7\% (AUC @ 10°) over the SuperEvent baseline, driven by our refined MCTS formulation. Since the evaluation enforces rotation changes of 1°--45°, the gains also indicate improved robustness to varying motion \emph{direction}, although our formulation explicitly normalizes only motion \emph{magnitude} and does not achieve full motion-invariance. As we employ the same training strategy as SuperEvent~\cite{Bur25}, the event-inherent robustness to HDR and motion blur is preserved. This is demonstrated in the drone experiments in Sec.~\ref{sec:drone_exp}.\par
The inference performance of SuperEvent+ is comparable to the baseline, as the only architectural distinction is the 8-channel input layer instead of 10. Our speed-optimized SuperLitE similarly improves the AUC@10° score by \SI{15.0}{\%} over the baseline. However, its inference takes only \SI{2.5}{ms}; \SI{90.5}{\%} faster than the baseline. This enables low-latency state estimation on compute-restricted onboard devices.\par
Fig.~\ref{se_infer} shows the better adaptability of our MCTS$_{Ne}$ formulation for motion speed changes. Even though both samples have large visual overlap, the MCTS$_{\Delta t}$ with constant time windows of SuperEvent result in highly different time surfaces in the \SI{10}{ms} channel. In contrast, the constant event count MCTS$_{Ne}$ channel with \SI{0.3}{events \per pixel} is visually more similar. This allows SuperEvent+ and SuperLitE to predict more correct matches that cover a larger area.


\begin{table*}[t]
\begin{center}
\caption{Root mean square (RMS) absolute trajectory error (ATE) [cm], median of 5 runs. Sequences where a system crashed or diverged in at least 3 runs are marked as ``failed''. Numbers in brackets correspond to the cropped sequences evaluated in~\cite{Niu24_b, Zho25}. ESVO2, SDEVO and OKVIS2-SE+ are evaluated on a desktop PC, all others on a Jetson Orin NX. OKVIS2-SE+ is the event-based version of OKVIS2 with our integrated SuperEvent+. AERO-VIS is our asynchronous extension using the faster SuperLitE. OKVIS2-SuperLitE (OKVIS2-SL) employs all performance optimizations of AERO-VIS except for asynchronous processing. For offline processing with OKVIS2-SE+ and OKVIS2-SL, the synchronous rates are \SI{50}{Hz} (small-scale: rpg-stereo, TUM-VIE) and \SI{20}{Hz} (large-scale: VECtor). We downsample the event streams to $240 \times 424$ pixels (TUM-VIE) and $240 \times 320$ (VECtor) for the Jetson evaluations.~\label{tab:data}}
\begin{tabularx}{\linewidth}{
l
l
>{\centering}X
>{\centering}X
>{\centering}X
>{\centering}X
>{\centering}X
>{\centering}X
>{\centering\arraybackslash}X
}
\toprule
& & \multicolumn{4}{c}{\textbf{Offline processing}} & \multicolumn{3}{c}{\textbf{Real-time on NVIDIA Jetson Orin NX}} \\
\cmidrule(lr){3-6}
\cmidrule(lr){7-9}
Dataset & Sequence & ESVO2~\cite{Niu24_b} & SDEVO~\cite{Zho25} & \textbf{OKVIS2-SE+ (ours)} & \textit{OKVIS2-SL (ablation)} & ESVO2~\cite{Niu24_b} & SDEVO~\cite{Zho25} & \textbf{AERO-VIS (ours)} \\

\cmidrule(lr){1-2}
\cmidrule(lr){3-6}
\cmidrule(lr){7-9}

\multirow{5}{*}{rpg-stereo}
 & bin & 3.73 \textcolor{gray}{(1.63)} & 3.78 \textcolor{gray}{(4.64)} & \textbf{0.56} \textcolor{gray}{(\textbf{0.41})} & \underline{1.44} \textcolor{gray}{(\underline{1.30})} & 216.71 & \underline{4.27} & \textbf{1.43} \\
 & boxes2 & 8.65 \textcolor{gray}{(3.57)} & 3.50 \textcolor{gray}{(\textbf{0.99})} & \textbf{1.60} \textcolor{gray}{(\underline{1.08})} & \underline{2.62} \textcolor{gray}{(3.03)} & 734.37 & \underline{23.14} & \textbf{2.86} \\
 & desk2 & 44.42 \textcolor{gray}{(3.27)} & 5.56 \textcolor{gray}{(\underline{0.59})} & \textbf{0.81} \textcolor{gray}{(\textbf{0.35})} & \underline{1.49} \textcolor{gray}{(1.67)} & 61.98 & \underline{27.46} & \textbf{1.26} \\
 & monitor2 & 1.87 \textcolor{gray}{(1.80)} & 2.73 \textcolor{gray}{(2.52)} & \textbf{0.60} \textcolor{gray}{(\textbf{0.50})} & \underline{1.61} \textcolor{gray}{(\underline{1.77})} & \underline{2.33} & 3.08 & \textbf{1.30} \\
 & reader & 123.82 \textcolor{gray}{(1.80)} & \underline{1.59} \textcolor{gray}{(\underline{0.53})} & \textbf{0.83} \textcolor{gray}{(\textbf{0.48})} & 3.24 \textcolor{gray}{(1.16)} & 102.02 & \underline{6.44} & \textbf{2.31} \\

\cmidrule(lr){1-2}
\cmidrule(lr){3-6}
\cmidrule(lr){7-9}

\multirow{7}{*}{TUM-VIE} 
 & 1d-trans & 2.19 & 1.10 & \textbf{0.34} & \underline{0.94} & failed & \underline{1.82} & \textbf{1.48} \\
 & 3d-trans & 14.45 & 1.39 & \textbf{0.65} & \underline{1.25} & failed & \underline{23.04} & \textbf{0.89} \\
 & 6dof & 3.48 & 2.22 & \textbf{0.38} & \underline{1.57} & failed & \underline{4.38} & \textbf{1.05} \\
 & desk & 9.14 & 2.52 & \textbf{0.36} & \underline{1.38} & failed & \underline{18.56} & \textbf{1.18} \\
 & desk2 & 8.73 & 2.06 & \textbf{0.31} & \underline{1.22} & failed & \underline{18.47} & \textbf{0.71} \\
 & shake & failed & 76.55 & \textbf{3.79} & \underline{8.61} & failed & \underline{91.61} & \textbf{8.13} \\
 & shake2 & failed & 59.09 & \textbf{9.15} & \underline{12.35} & failed & \underline{73.64} & \textbf{13.46} \\

\cmidrule(lr){1-2}
\cmidrule(lr){3-6}
\cmidrule(lr){7-9}

\multirow{6}{*}{VECtor}
 & corr.-dolly & failed & 126.61 & \textbf{73.28} & \underline{110.15} & failed & \underline{352.92} & \textbf{174.55} \\
 & corr.-walk & failed & 331.33 & \textbf{70.22} & \underline{259.03} & failed & \underline{290.12} & \textbf{237.36} \\
 & school-dolly & 1448.08 & \underline{101.46} & \textbf{74.51} & 527.54 & failed & \textbf{247.86} & \underline{281.45} \\
 & school-scooter & failed & 824.45 & \textbf{174.44} & \underline{371.51} & failed & \underline{612.34} & \textbf{256.20} \\
 & units-dolly & failed & 474.20 & \textbf{136.72} & \underline{236.51} & failed & \underline{564.34} & \textbf{207.08} \\
 & units-scooter & failed & 1530.22 & \textbf{138.48} & \underline{393.19} & failed & \underline{2828.34} & \textbf{326.76} \\
 
\bottomrule
\end{tabularx}
\vspace{-11pt}
\end{center}
\end{table*}

\afterpage{
\afterpage{
\begin{figure}[t]
  \vspace{-4pt}
  \includegraphics[width=\columnwidth]{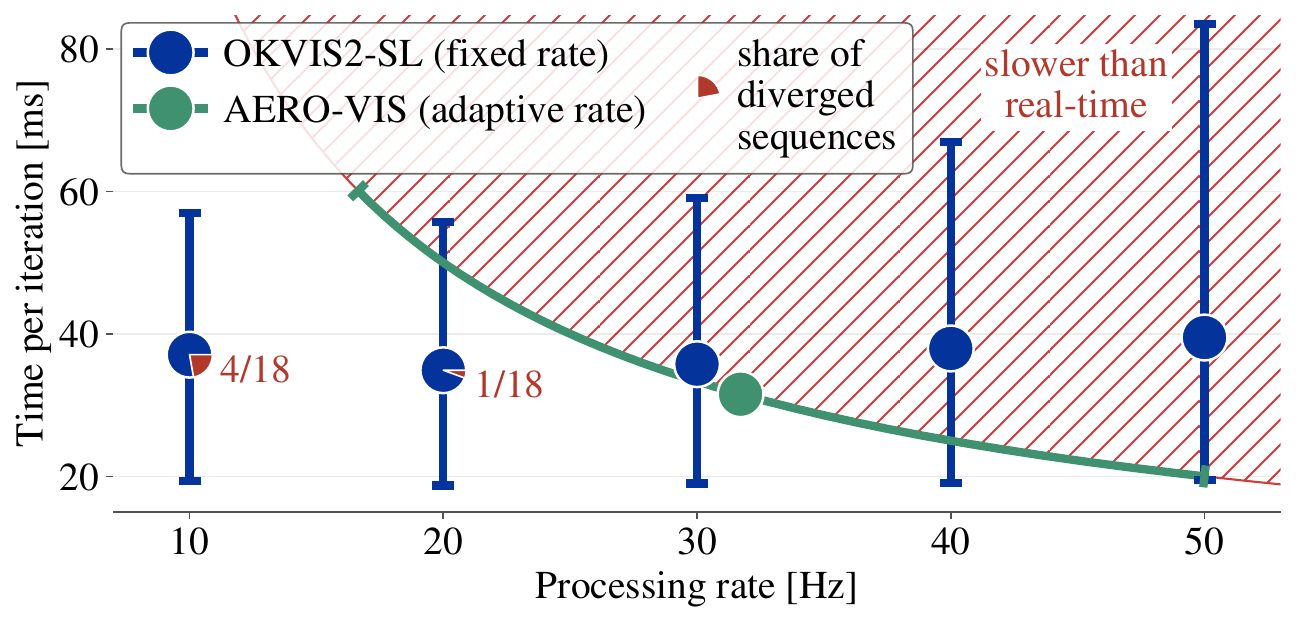}
  \vspace{-16pt}
  \caption{Ablation study of system throughput for 5 synchronous frequencies (OKVIS2-SL) and asynchronous (AERO-VIS) processing, both run with full performance optimization. Bars show the 1st–99th percentiles of the distributions of iteration times and processing rates, dots mark the mean, across all evaluation sequences. AERO-VIS runs continuously at the real-time boundary since its rate is set directly by the per-iteration compute time. Sequences with ATE exceeding \SI{10}{\%} of their total length are considered as \textit{diverged} (annotated in red) and excluded from the timing statistics.}
  \label{throughput}
  \vspace{-10pt}
\end{figure}
}
}

\subsection{Dataset Evaluation}~\label{sec:data}
We benchmark our proposed system on three public datasets. The rpg-stereo~\cite{Zho18} dataset provides small-scale trajectories recorded with two iniVation DAVIS240C cameras. We found a temporal misalignment between the sensor data and pose ground truth. Therefore, we offline estimate a time offset within $\pm$\SI{100}{ms} which minimizes the error for all estimators. The second dataset, TUM-VIE~\cite{Kle21}, was recorded with Prophesee Gen4 HD cameras at a $720\times1280$ pixel resolution. Only its small-scale \textit{mocap}-sequences have consistent ground truth poses. The \textit{mocap-shake} sequences exhibit aggressive camera shaking and partially featureless views, making estimation highly challenging. Finally, we benchmark on the large-scale sequences of the VECtor~\cite{Gao22} dataset. It employs Prophesee Gen3 CD cameras with $480\times640$ pixels. The indoor sequences contain few and similar features. Vignetting artifacts from the infrared (IR) filter and an imprecise extrinsic calibration of the event cameras introduce additional challenges. Since we were unable to obtain a set of calibration parameters that reliably work on all sequences, we enable the online extrinsic calibration inherited from OKVIS2.\par
Our SLAM framework offers configurable operating modes: an accuracy-optimized setting for post-processing recorded data, and a performance-oriented setting for real-time applications. For maximum accuracy in offline processing, we employ the full SuperEvent+ network integrated into OKVIS2 with all runtime-optimizations disabled -- including asynchronous processing, which is imposed by real-time constraints. Instead, the synchronous processing rate is set to 50 Hz for small-scale datasets (rpg-stereo and TUM-VIE) and 20 Hz for the large-scale sequences in VECtor. We use a desktop PC with an Intel Core i5-13600 processor, an NVIDIA GeForce RTX 4070 GPU, and 32 GB RAM.\par

\afterpage{
\afterpage{
\begin{table}[t]
\begin{center}
\caption{RMS of ATE [cm] for indoor drone experiments.\label{tab:drone}}
\begin{tabularx}{\linewidth}{
l
l
l
r
r
}
\toprule
Type of & Lighting & Control & OKVIS2~\cite{Leu22} & \textbf{AERO-} \\
movement & condition & mode & (frame-based) & \textbf{VIS} \\
\midrule
``8'' shape & normal & closed-loop & \textbf{2.14} & 10.83 \\

In-place rot. & HDR & closed-loop & failed$^*$ & \textbf{25.53} \\

Aggressive & normal & handheld & 78.08 & \textbf{7.78} \\
\bottomrule
\vspace{-6pt}
\end{tabularx}
\footnotesize{
$^*$Drone crashed after large drift.
}
\vspace{-16pt}
\end{center}
\end{table}
}
}

To validate AERO-VIS's robustness for onboard applications, we evaluate it on an NVIDIA Jetson Orin NX with enforced asynchronous real-time processing. We employ the performance-optimized SuperLitE quantized to 16-bit floating-point precision, compiled by TensorRT, and with CUDA graph capture. Additionally, we limit the backend optimization's maximum iterations and time budget. We spatially downsample the event stream to a lower resolution to align with the hardware’s real-time processing bandwidth. Finally, we evaluate OKVIS2-SuperLitE with full performance-optimization at the synchronous frequencies of OKVIS2-SuperEvent+ to ablate our proposed system architecture.\par
In Table~\ref{tab:data}, we compare the results of our two system variants with the two SOTA baselines: the stereo event-inertial ESVO2~\cite{Niu24_b} and the stereo event-based SDEVO~\cite{Zho25}. Both report real-time capability (on a Desktop PC) in their publications. For a fair comparison, we evaluate the baselines on the same hardware and at the same spatial resolutions as our systems. While \cite{Tej26} is also relevant for our evaluation, its source code has not been released publicly, and the system was not benchmarked on public datasets.\par
Even without real-time constraints, ESVO2 fails in scenes with high optical flow (e.g., TUM-VIE mocap-shake \& -shake2, and VECtor sequences) but also has inconsistent results on the other sequences. The second baseline, SDEVO, performs better on most sequences while having large errors on others (e.g., rpg-stereo desk2, TUM-VIE mocap-shake \& -shake2, VECtor school-scooter \& units-scooter). This lack of robustness could be caused by a domain shift compared to its training data or the missing integration of IMU measurements. OKVIS2-SuperEvent+ achieves the best results on all sequences. This can be attributed to the reliable keypoint detection and description of SuperEvent+ as well as OKVIS2's sensor fusion and backend optimization.\par
In the real-time evaluations on the Jetson Orin NX, ESVO2 exhibits significant instability, characterized by failure or severe drift across most sequences. These results suggest its computational demands exceed the capabilities of the onboard hardware. The accuracy of SDEVO is constrained by its low inference speed (1–\SI{3}{Hz}), leading to substantial estimation errors across most test sequences. AERO-VIS, however, achieves precise results and -- on most sequences -- even outperforms ESVO2 and SDEVO's unconstrained desktop results. Crucially, AERO-VIS is the \textit{only} evaluated system capable of reliable onboard state estimation.\par
To ablate our asynchronous system design, we compare AERO-VIS with OKVIS2-SuperLitE in terms of accuracy and performance. AERO-VIS's accuracy results in Table~\ref{tab:data} are comparable with or better than those of the synchronous system. Its adaptive rate provides fast enough processing without flooding the backend with observations that would impair precise estimation within the restricted optimization time budget. Additionally, Fig.~\ref{throughput} compares the performance of AERO-VIS and OKVIS2-SuperLitE at different processing rates, demonstrating that finding a suitable synchronous frequency for real-time operation is not trivial since it depends on the data (e.g., number of events, number of detected keypoints, complexity of the optimization problem, etc.). Even when the mean processing time is within the real-time budget, many samples can still exceed the real-time boundary, e.g., at 20 Hz. At the same time, samples below the boundary limit throughput, as the system must wait for their arrival. Also, low processing rates impact robustness as more sequences diverge. The asynchronous system design of AERO-VIS operates at the fastest possible instantaneous rate, thereby maximizing throughput while strictly adhering to the real-time budget.


\subsection{UAV Onboard Estimation}~\label{sec:drone_exp}
To validate AERO-VIS's performance in real-world scenarios and its potential for applications such as drone control, we employ the system on a custom-built UAV. The platform features an NVIDIA Jetson Orin NX, two Prophesee EVK4 event cameras, and a Bosch BMI160 IMU. The stereo configuration utilizes a \SI{7.4}{cm} baseline and 110° FOV lenses, equipped with IR filters to prevent interference with the motion-capture system. To allow robust online event processing, we spatially downsample the event stream to $240\times424$ pixels on the hardware and additionally restrict the maximum event rate to \SI{35}{MEv \per s}. As a drone controller, we use a linear MPC~\cite{Tzo19} to follow predefined trajectories.\par
We evaluate three scenarios. Firstly, we employ AERO-VIS with closed-loop control in a well-lit environment executing an ``8''-shaped trajectory five times, covering an area of $\sim$\SI{5}{m}$^2$. Secondly, we turn off the lights and place a light source behind a window to test the system's robustness to HDR. Again, the UAV is controlled by AERO-VIS's real-time estimation to spin 10 times in place. Finally, we simulate aggressive flight with a handheld shake of the drone in the horizontal plane and rapid rotation around its vertical axis. We measure ground-truth trajectories using a motion-capture system to evaluate our system's accuracy. As a baseline, the original OKVIS2~\cite{Leu22} is evaluated in the same experiments using the frame camera RealSense 455 with a $480\times640$ pixel resolution and the integrated Bosch BMI055 IMU.\par
Table~\ref{tab:drone} lists the resulting trajectory estimation errors in the different experiments, and Fig.~\ref{drone_traj} shows the estimated and measured trajectories. Under normal conditions, OKVIS2 outperforms AERO-VIS. The precision of AERO-VIS is limited by reduced spatial resolution and sensitivity of time surfaces to vibration-induced appearance changes. The adopted training strategy~\cite{Bur25} relies on consecutive samples from the same sequence, limiting network exposure to sudden motion variations. Consequently, because vibrations continuously shift motion direction and distort feature appearance within the event stream, event-based descriptor matching degrades under these dynamics compared to standard frames. While our MCTS formulation mitigates this effect as shown in Section~\ref{sec:se}, it does not eliminate it entirely. However, the other two experiments demonstrate the event camera's increased robustness to HDR and rapid motion. In the HDR experiment, the estimate of AERO-VIS drifts by $\sim$\SI{1}{m}. We identified rotation-only trajectories as a constraint for OKVIS2’s loop closure detection. Since the system enforces a bound on the maximum relative linear drift, the lack of significant translational movement effectively prevents loop-closure triggers. Still, AERO-VIS demonstrates superior stability compared to the frame-based baseline. While both systems experience drift, OKVIS2's error magnitude is substantially higher, causing the UAV to crash into the wall after 5 rotations. In the third experiment simulating aggressive flight, AERO-VIS reduces the estimation error by \SI{90}{\%} compared to frame-based OKVIS2 due to the event camera's higher temporal resolution and lower sensitivity to motion blur, resulting in significantly less drift.


\begin{figure}[t]
  \includegraphics[width=\columnwidth]{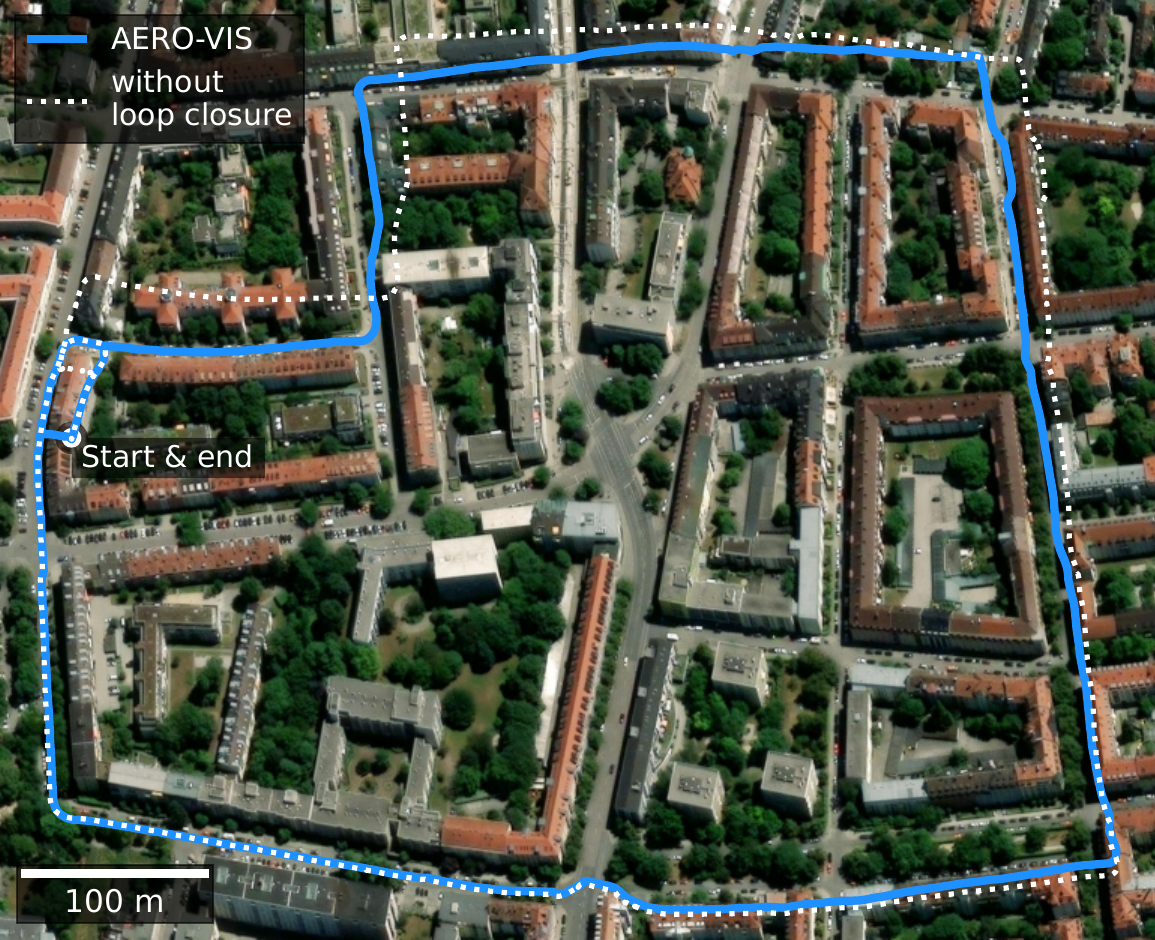}
  \caption{Estimated trajectory during a 2 km (20 min) urban walk, processed online on an NVIDIA Jetson Orin NX. The initial pose was manually aligned with the map.}
  \label{ls_traj}
\end{figure}

\afterpage{
\begin{table}[t]
\begin{center}
\caption{Mean cycle time of AERO-VIS's components during the large-scale real-world experiment [ms].\label{tab:timings}}
\begin{tabularx}{\linewidth}{
l
>{\centering}X
>{\centering}X
>{\centering}X
>{\centering\arraybackslash}X
}
\toprule
\textbf{Time interval [min]} & $\mathbf{0-5}$ & $\mathbf{5-10}$ & $\mathbf{10-15}$ & $\mathbf{15-20}$ \\
\midrule
MCTS Preparation$^{1,2}$ & 10.2 & 9.6 & 9.3 & 9.3 \\

Keypoint detection$^2$ & 14.3 & 13.7 & 13.4 & 13.2 \\

Matching \& filtering & 27.6 & 42.2 & 45.0 & 48.9 \\

Loop closure detection & 18.6 & 67.3 & 96.9 & 106.2 \\

Backend optimization$^2$ & 21.9 & 31.8 & 38.0 & 43.2 \\



Other$^3$ & 16.1 & 15.0 & 16.6 & 15.4 \\
\midrule
Total cycle time & 84.2 & 156.3 & 196.5 & 213.7 \\

\bottomrule
\vspace{-6pt}
\end{tabularx}
\footnotesize{
$^1$Runs in \SI{50}{Hz} real-time thread. \qquad $^2$Run in parallel.\\
$^3$Auxiliary tasks, inter-process communication, memory management.
}
\vspace{-18pt}
\end{center}
\end{table}
}

\subsection{Large-scale Real-world Loop}~\label{sec:ls}
Finally, we qualitatively analyze AERO-VIS in a large-scale real-world experiment. Using the same hardware setup from Sec.~\ref{sec:drone_exp}, we walk a loop of \SI{2}{km} in \SI{20}{min} in an urban environment containing various lighting conditions (sun, shade, artificial light) and dynamic entities (cars, pedestrians, bicycles). The online-estimated trajectory is overlaid on a satellite image in Fig.~\ref{ls_traj}. In the absence of ground truth, the qualitative accuracy of the system is evident through the close correspondence between the trajectory estimation and the actual street layout. Upon loop-closure detection, a trajectory drift of \SI{1.8}{\%} relative to the estimated traversed distance is corrected. This experiment further shows our system's robustness and its relevance to real-world robotics.\par
Table~\ref{tab:timings} lists the evolving timings of several system components. The per-iteration runtime of our proposed extensions -- MCTS preparation and the keypoint detection -- is nearly constant throughout the experiment. The keypoint detection timing combines SuperLitE inference, non-maximum suppression, descriptor interpolation, and descriptor quantization. 
As the trajectory length increases, the loop closure detection becomes a significant bottleneck. Keypoint matching and backend optimization also contribute to rising overhead over time. Since these limitations are inherent to the integrated OKVIS2 framework and persist in the original system's frame-based processing, we consider the optimization of these components to be outside the scope of this paper. However, even when processing rates drop below \SI{5}{Hz}, AERO-VIS reliably estimates the trajectory.

\section{Conclusion}
We presented AERO-VIS, an event-inertial SLAM system that integrates our performance-optimized keypoint detection and description network, SuperLitE, into our asynchronous version of OKVIS2 for onboard event processing. Even with severely restricted computational resources, it outperforms current SOTA results obtained without these restrictions.\par
We conduct several real-world experiments. Firstly, we demonstrate closed-loop UAV control by onboard event-based state estimation. Secondly, AERO-VIS outperforms the original frame-based OKVIS2 in HDR and aggressive motion scenarios. Finally, our system produces accurate estimates over a long time horizon and reliably detects loop closures.\par
While our method reduces the motion-dependence of event-based descriptors, they remain less motion-invariant than their frame-based counterparts in the presence of UAV vibrations, leaving a performance gap between AERO-VIS and OKVIS2 under standard flight conditions. By efficiently merging event-based keypoint detections with other modalities, such as frames or LiDAR, a hybrid system could potentially achieve the high accuracy of OKVIS2 alongside the robustness to HDR and fast motion of AERO-VIS.\par
Ultimately, AERO-VIS demonstrates that event-inertial SLAM can meet the demands of onboard UAV control. By addressing the trade-off between low-latency sensing and estimation precision, this work provides a viable path toward autonomous navigation where traditional vision fails.

\section*{Acknowledgment}
We thank Rhythm Chandak for her efforts in hardware integration, as well as Cédric Le Gentil and Leonard Diederich for providing valuable feedback on the manuscript.



\renewcommand{\baselinestretch}{0.969} 
\bibliographystyle{bib/IEEEtran.bst}
\bibliography{bib/IEEEabrv.bib, bib/9_bib}

\begin{thebibliography}{10}
\providecommand{\url}[1]{#1}
\csname url@rmstyle\endcsname
\providecommand{\newblock}{\relax}
\providecommand{\bibinfo}[2]{#2}
\providecommand\BIBentrySTDinterwordspacing{\spaceskip=0pt\relax}
\providecommand\BIBentryALTinterwordstretchfactor{4}
\providecommand\BIBentryALTinterwordspacing{\spaceskip=\fontdimen2\font plus
\BIBentryALTinterwordstretchfactor\fontdimen3\font minus \fontdimen4\font\relax}
\providecommand\BIBforeignlanguage[2]{{%
\expandafter\ifx\csname l@#1\endcsname\relax
\typeout{** WARNING: IEEEtran.bst: No hyphenation pattern has been}%
\typeout{** loaded for the language `#1'. Using the pattern for}%
\typeout{** the default language instead.}%
\else
\language=\csname l@#1\endcsname
\fi
#2}}

\bibitem{Bur25}
Y.~Burkhardt, S.~Schaefer, and S.~Leutenegger, ``{SuperEvent}: Cross-modal learning of event-based keypoint detection for {SLAM},'' in \emph{IEEE/CVF Int. Conf. Comput. Vis. (ICCV)}, 2025.

\bibitem{Niu24_b}
J.~Niu, S.~Zhong, X.~Lu, S.~Shen, G.~Gallego, and Y.~Zhou, ``{ESVO2}: Direct visual-inertial odometry with stereo event cameras,'' \emph{IEEE Trans. Robot.}, 2025.

\bibitem{Zho25}
S.~Zhong, J.~Niu, and Y.~Zhou, ``Deep visual odometry for stereo event cameras,'' \emph{IEEE Robot. Automat. Lett.}, 2025.

\bibitem{Che23}
P.~Chen, W.~Guan, and P.~Lu, ``{ESVIO}: Event-based stereo visual inertial odometry,'' \emph{IEEE Robot. Automat. Lett.}, 2023.

\bibitem{Tej26}
D.~Tejero-Ruiz, D.~Sol{\'\i}s-Mart{\'\i}n, F.~J. P{\'e}rez-Grau, and J.~Borrego-D{\'\i}az, ``Indoor {UAV} navigation using event cameras and intermediate frame reconstruction,'' \emph{Comput. Vis. Image Understand.}, 2026.

\bibitem{Leu22}
S.~Leutenegger, ``{OKVIS2}: Realtime scalable visual-inertial {SLAM} with loop closure,'' \emph{arXiv preprint}, 2022.

\bibitem{Zhu17}
A.~Z. Zhu, N.~Atanasov, and K.~Daniilidis, ``Event-based visual inertial odometry,'' in \emph{IEEE/CVF Conf. Comput. Vis. Pattern Recog. (CVPR)}, 2017.

\bibitem{Had21}
A.~Hadviger, I.~Cvišić, I.~Marković, S.~Vražić, and I.~Petrović, ``Feature-based event stereo visual odometry,'' in \emph{Eur. Conf. Mobile Robots (ECMR)}, 2021.

\bibitem{Lag17}
X.~Lagorce, G.~Orchard, F.~Galluppi, B.~E. Shi, and R.~B. Benosman, ``{HOTS}: A hierarchy of event-based time-surfaces for pattern recognition,'' \emph{{IEEE} Trans. Pattern Anal. Machine Intell.}, 2017.

\bibitem{Alz18}
I.~Alzugaray and M.~Chli, ``Asynchronous corner detection and tracking for event cameras in real time,'' \emph{IEEE Robot. Automat. Lett.}, 2018.

\bibitem{Cha22}
W.~Chamorro, J.~Sola, and J.~Andrade-Cetto, ``Event-based line slam in real-time,'' \emph{IEEE Robot. Automat. Lett.}, 2022.

\bibitem{Reb18}
H.~Rebecq, G.~Gallego, E.~Mueggler, and D.~Scaramuzza, ``{EMVS}: Event-based multi-view stereo—{3D} reconstruction with an event camera in real-time,'' \emph{Int. J. Comput. Vis.}, 2018.

\bibitem{Gua22}
W.~Guan and P.~Lu, ``Monocular event visual inertial odometry based on event-corner using sliding windows graph-based optimization,'' in \emph{IEEE/RSJ Int. Conf. Intell. Robots Syst. (IROS)}, 2022.

\bibitem{Kan81}
B.~D. Lucas and T.~Kanade, ``An iterative image registration technique with an application to stereo vision,'' in \emph{Int. Joint Conf. Artif. Intell. (IJCAI)}, 1981.

\bibitem{Kim16}
H.~Kim, S.~Leutenegger, and A.~J. Davison, ``Real-time {3D} reconstruction and 6-dof tracking with an event camera,'' in \emph{Eur. Conf. Comput. Vis. (ECCV)}, 2016.

\bibitem{Reb16}
H.~Rebecq, T.~Horstsch{\"a}fer, G.~Gallego, and D.~Scaramuzza, ``{EVO}: A geometric approach to event-based 6-dof parallel tracking and mapping in real time,'' \emph{IEEE Robot. Automat. Lett.}, 2016.

\bibitem{Gho24}
S.~Ghosh, V.~Cavinato, and G.~Gallego, ``{ES-PTAM}: Event-based stereo parallel tracking and mapping,'' in \emph{Eur. Conf. Comput. Vis. Workshops (ECCVW)}, 2024.

\bibitem{Gho22}
S.~Ghosh and G.~Gallego, ``Multi-event-camera depth estimation and outlier rejection by refocused events fusion,'' \emph{Adv. Intell. Syst.}, 2022.

\bibitem{Zho21}
Y.~Zhou, G.~Gallego, and S.~Shen, ``Event-based stereo visual odometry,'' \emph{IEEE Trans. Robot.}, 2021.

\bibitem{Liu23_t}
Z.~Liu, D.~Shi, R.~Li, Y.~Zhang, and S.~Yang, ``{T-ESVO}: improved event-based stereo visual odometry via adaptive time-surface and truncated signed distance function,'' \emph{Adv. Intell. Syst.}, 2023.

\bibitem{Liu23}
Z.~Liu, D.~Shi, R.~Li, and S.~Yang, ``{ESVIO}: event-based stereo visual-inertial odometry,'' \emph{Sensors}, 2023.

\bibitem{Niu24_a}
J.~Niu, S.~Zhong, and Y.~Zhou, ``{IMU}-aided event-based stereo visual odometry,'' in \emph{IEEE Int. Conf. Robot. Automat. (ICRA)}, 2024.

\bibitem{Ye18}
C.~Ye, A.~Mitrokhin, C.~Ferm{\"u}ller, J.~A. Yorke, and Y.~Aloimonos, ``Unsupervised learning of dense optical flow, depth and egomotion from sparse event data,'' \emph{arXiv preprint}, 2018.

\bibitem{Zhu19}
A.~Z. Zhu, L.~Yuan, K.~Chaney, and K.~Daniilidis, ``Unsupervised event-based learning of optical flow, depth, and egomotion,'' in \emph{IEEE/CVF Conf. Comput. Vis. Pattern Recog. (CVPR)}, 2019.

\bibitem{Kle24}
S.~Klenk, M.~Motzet, L.~Koestler, and D.~Cremers, ``Deep event visual odometry,'' in \emph{Int. Conf. 3D Vis. (3DV)}, 2024.

\bibitem{Tee23}
Z.~Teed, L.~Lipson, and J.~Deng, ``Deep patch visual odometry,'' \emph{Adv. Neural Inf. Process. Syst. (NeurIPS)}, 2023.

\bibitem{Gua24_b}
W.~Guan, F.~Lin, P.~Chen, and P.~Lu, ``{DEIO}: Deep event inertial odometry,'' in \emph{IEEE/CVF Int. Conf. Comput. Vis. Workshops (ICCVW)}, 2025.

\bibitem{Sch20}
C.~Scheerlinck, H.~Rebecq, D.~Gehrig, N.~Barnes, R.~Mahony, and D.~Scaramuzza, ``Fast image reconstruction with an event camera,'' in \emph{IEEE/CVF Winter Conf. Appl. Comput. Vis. (WACV)}, 2020.

\bibitem{Gen20}
P.~Geneva, K.~Eckenhoff, W.~Lee, Y.~Yang, and G.~Huang, ``{OpenVINS}: A research platform for visual-inertial estimation,'' in \emph{IEEE Int. Conf. Robot. Automat. (ICRA)}, 2020.

\bibitem{Par24}
F.~Paredes-Vallés, J.~J. Hagenaars, J.~Dupeyroux, S.~Stroobants, Y.~Xu, and G.~C. H.~E. de~Croon, ``Fully neuromorphic vision and control for autonomous drone flight,'' \emph{Sci. Robot.}, 2024.

\bibitem{Sek26}
Y.~Sekikawa, J.~Nagata, I.~Araki, and A.~Girbau, ``{CoL2A}: Convolution-free local linear attention for spatiotemporal event processing,'' in \emph{IEEE/CVF Winter Conf. Appl. Comput. Vis. (WACV)}, 2026.

\bibitem{Hao26}
H.~Hao, Z.~Sui, R.~Zou, Z.~Dai, N.~Zubi{\'c}, D.~Scaramuzza, and W.~Wang, ``Low-latency event-based object detection with spatially-sparse linear attention,'' \emph{arXiv preprint}, 2026.

\bibitem{Det18}
D.~DeTone, T.~Malisiewicz, and A.~Rabinovich, ``{SuperPoint}: Self-supervised interest point detection and description,'' in \emph{IEEE/CVF Conf. Comput. Vis. Pattern Recog. Workshops (CVPRW)}, 2018.

\bibitem{Tu22}
Z.~Tu, H.~Talebi, H.~Zhang, F.~Yang, P.~Milanfar, A.~Bovik, and Y.~Li, ``{MaxViT}: Multi-axis vision transformer,'' in \emph{Eur. Conf. Comput. Vis. (ECCV)}, 2022.

\bibitem{Sim14}
K.~Simonyan, ``Very deep convolutional networks for large-scale image recognition,'' \emph{arXiv preprint}, 2014.

\bibitem{How19}
A.~Howard, M.~Sandler, G.~Chu, L.-C. Chen, B.~Chen, M.~Tan, W.~Wang, Y.~Zhu, R.~Pang, V.~Vasudevan, Q.~V. Le, and H.~Adam, ``Searching for {MobileNetV3},'' in \emph{IEEE/CVF Int. Conf. Comput. Vis. (ICCV)}, 2019.

\bibitem{Gal12}
D.~G{\'a}lvez-L{\'o}pez and J.~D. Tardos, ``Bags of binary words for fast place recognition in image sequences,'' \emph{IEEE Trans. Robot.}, 2012.

\bibitem{Mue17_ecd}
E.~Mueggler, H.~Rebecq, G.~Gallego, T.~Delbruck, and D.~Scaramuzza, ``The event-camera dataset and simulator: Event-based data for pose estimation, visual odometry, and slam,'' \emph{Int. J. Robot. Res.}, 2017.

\bibitem{Hid22}
J.~Hidalgo-Carri{\'o}, G.~Gallego, and D.~Scaramuzza, ``Event-aided direct sparse odometry,'' in \emph{IEEE/CVF Conf. Comput. Vis. Pattern Recog. (CVPR)}, 2022.

\bibitem{Zho18}
Y.~Zhou, G.~Gallego, H.~Rebecq, L.~Kneip, H.~Li, and D.~Scaramuzza, ``Semi-dense {3D} reconstruction with a stereo event camera,'' in \emph{Eur. Conf. Comput. Vis. (ECCV)}, 2018.

\bibitem{Kle21}
S.~Klenk, J.~Chui, N.~Demmel, and D.~Cremers, ``{TUM-VIE}: The {TUM} stereo visual-inertial event dataset,'' in \emph{IEEE/RSJ Int. Conf. Intell. Robots Syst. (IROS)}, 2021.

\bibitem{Gao22}
L.~Gao, Y.~Liang, J.~Yang, S.~Wu, C.~Wang, J.~Chen, and L.~Kneip, ``{VECtor}: A versatile event-centric benchmark for multi-sensor {SLAM},'' \emph{IEEE Robot. Automat. Lett.}, 2022.

\bibitem{Tzo19}
D.~Tzoumanikas, W.~Li, M.~Grimm, K.~Zhang, M.~Kovac, and S.~Leutenegger, ``Fully autonomous micro air vehicle flight and landing on a moving target using visual--inertial estimation and model-predictive control,'' \emph{J. Field Robot.}, 2019.

\end{thebibliography}

\end{document}